\documentclass{article}

\usepackage{graphicx}

\usepackage{amsmath}
\usepackage[utf8]{inputenc} 
\usepackage[T1]{fontenc}    
\usepackage{hyperref}       
\usepackage{url}            
\usepackage{booktabs}       
\usepackage{amsfonts}       
\usepackage{nicefrac}       
\usepackage{microtype}      
\usepackage{xcolor}         
\usepackage{subcaption}
\usepackage{amsmath}
\usepackage[nonatbib]{neurips_2025}

\title{MetaCD: A Meta Learning Framework for Cognitive Diagnosis based on Continual Learning}

\begin{document}
\maketitle

\begin{abstract}
Cognitive diagnosis is an essential research topic in intelligent education, aimed at assessing the level of mastery of different skills by students. So far, many research works have used deep learning models to explore the complex interactions between students, questions, and skills. However, the performance of existing method is frequently limited by the long-tailed distribution and dynamic changes in the data. To address these challenges, we propose a meta-learning framework for cognitive diagnosis based on continual learning (MetaCD). This framework can alleviate the long-tailed problem by utilizing meta-learning to learn the optimal initialization state, enabling the model to achieve good accuracy on new tasks with only a small amount of data. In addition, we utilize a continual learning method named parameter protection mechanism to give MetaCD the ability to adapt to new skills or new tasks, in order to adapt to dynamic changes in data. MetaCD can not only improve the plasticity of our model on a single task, but also ensure the stability and generalization of the model on sequential tasks. Comprehensive experiments on five real-world datasets show that MetaCD outperforms other baselines in both accuracy and generalization.
\end{abstract}

\section{Introduction}
\label{sec:introduction}
\noindent Cognitive diagnosis is a crucial educational measurement model in intelligent education, which aims to explore students' cognitive processes in problem-solving through their measurement data \cite{su2024constructing}. In general, the cognitive diagnosis system can assesses students' mastery of various skills by modeling cognitive processing. And it can provide timely feedback on students' weak skills \cite{akata2025playing}. 

Unfortunately, with the explosive growth of educational data, traditional cognitive diagnostic models (such as IRT \cite{edelen2007applying}  and DINA \cite{de2009dina}) cannot mine the potential nonlinear interactive relationships between students and questions \cite{gandhi2023understanding,tessler2024ai}, which limits models' comprehensive understanding of students' cognitive processes. In order to address these limitations, recent efforts has begun exploring cognitive diagnosis models based on deep learning, such as neural network-based cognitive diagnosis (NCD) \cite{wang2020neural}, self-supervised graph neural network-based  cognitive diagnosis (SCD) \cite{wang2023self}, etc. These approaches can mine the deep non-linear interaction between students-questions-skills.  

However, applying existing cognitive diagnostic methods to online learning systems still faces numerous challenges: \textit{\textbf{(a)}} some questions receive very few responses from students, while others receive many, resulting in sparse data and subsequently leading to long-tailed problems \cite{zhang2023deep}. \textit{\textbf{(b)}} student data in online learning systems typically exhibit two dynamic changes: (1) students learn new skills over time, and their mastery of previously learned skills changes; (2) new learning tasks appear  in the system can lead to changes in the data. Existing methods, designed for fixed datasets \cite{wang2020neural,frasson2021enhancing,su2022graph,qi2023icd}, struggle to adapt to these dynamic conditions. They typically exhibit significant performance degradation on new, emerging data patterns due to an inability to integrate new knowledge effectively. Furthermore, they are highly susceptible to catastrophic forgetting, whereby learning new information causes a rapid and severe loss of previously acquired knowledge.

In order to solve the above problems, we proposed a meta learning framework for cognitive diagnosis based on continual learning, called MetaCD. Our approach focuses more on considering the model's plasticity and stability in adapting to the long-tailed distribution, and dynamic changes in the data and new tasks. Our key contributions are summarized as follows: \textbf{\textit{(1)}}. We empower the knowledge base module through meta-learning methods to ensure that the model can alleviate the problem of long-tailed distribution through few data, achieving the reliability and robustness of MetaCD on sparse data. \textbf{\textit{(2)}}. We use parameter protection mechanism (PPM) to ensure the stability and plasticity of the model, enabling MetaCD to adapt to the dynamic changes of data and achieve its continual learning ability. \textbf{\textit{(3)}}. We use a knowledge extraction method based on Kullback-Leibler divergence to avoid fuzzy boundaries of diagnostic results, thereby improving the classification accuracy of the model. \textbf{\textit{(4)}}. We conduct comprehensive experiments on five real-world datasets to validate the performance of MetaCD, particularly on long-tailed data and task incremental learning.

\section{Our Proposed MetaCD Method}

\begin{figure*}
    \centering 
    \includegraphics[width=1.0\textwidth]{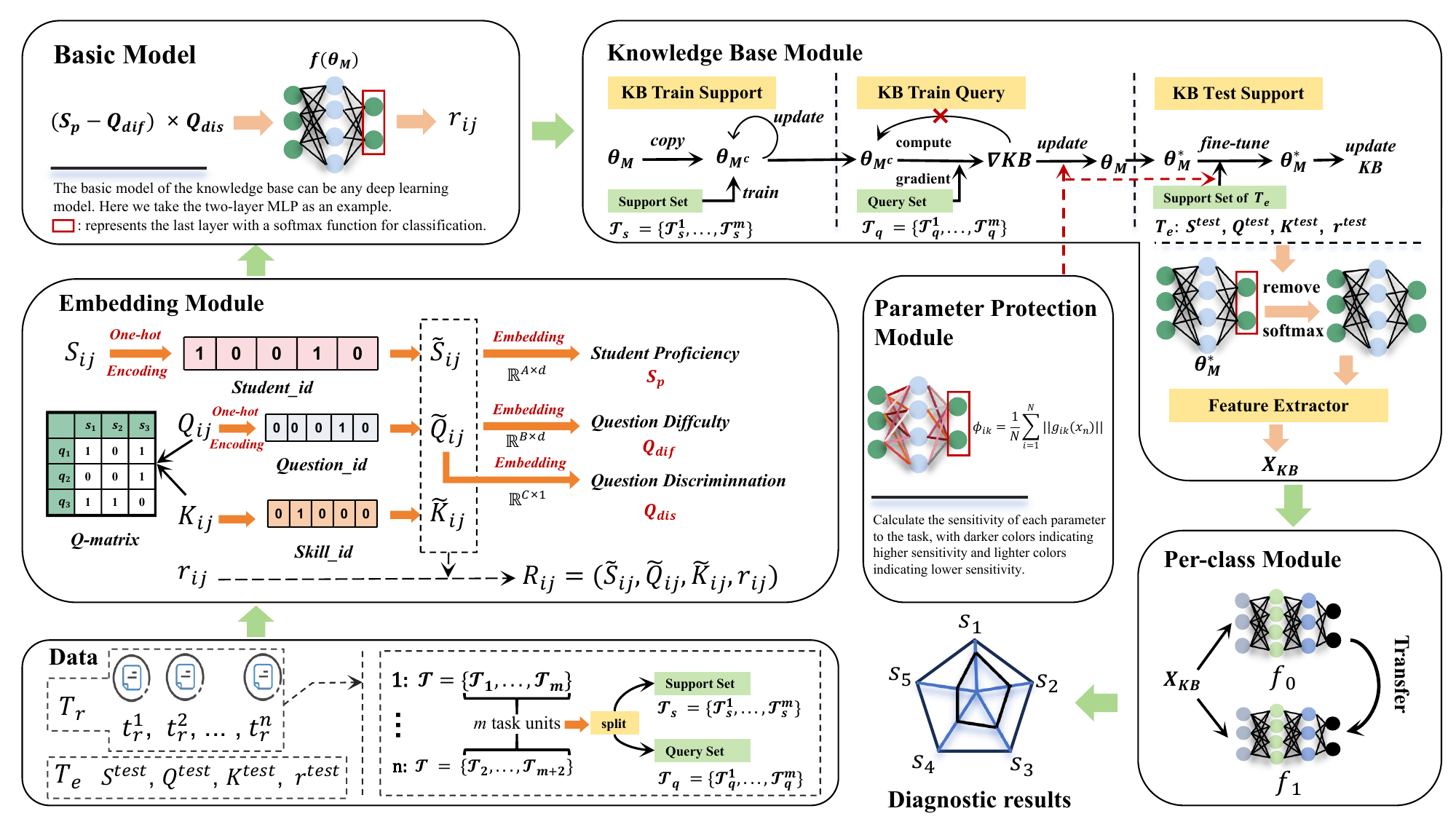}  
    \caption{The overall structure of MetaCD.} 
    \label{fig1}
\end{figure*}

\subsection{Task Overview}  
\label{sec:task-overview}
\noindent Suppose that we are given $n$ training task units set $T_r = \{ t_r^{1}, t_r^{2}, ...,  t_r^{n} \}$ and a testing task $T_e = \{ S^{test}, Q^{test}, K^{test}, r^{test} \}$. The data of each task unit $t_r^{i}$ consists of response logs of students, including student IDs $S_i$, question IDs $Q_i$, and skill IDs $K_i$, $i \in [1: n]$. Each student has a corresponding score $s$ for each question,  $s \in \{0, 1\}$ (1 indicates that the student answered a question correctly, otherwise 0) and $s \in r_i$, so $T_i = \{ S_i, Q_i,  K_i, r_i\}$. In addition, we need to construct the $\mathcal{Q}$-matrix\footnote{The $\mathcal{Q}$-matrix differs from $Q_i$. $Q_i$ represents the ID of the question answered by the students, while the $\mathcal{Q}$-matrix shows the binary relationship between questions and skills.} \cite{wang2020neural} for each task unit based on the questions and their corresponding skills. 

\noindent \textbf{Problem Definition:} Given a set of students' response data from training task units set $T_r$ and $\mathcal{Q}$-matrix, the purpose of MetaCD is to learn a model that can predict students' proficiency in skills and their corresponding scores by effectively transferring knowledge across tasks. The ultimate goal is to obtain the best model that can accurately predict students' performance on the testing task $T_e$. 

\subsection{The Structure of MetaCD}
\noindent {\hyperref[fig1]{Figure~1} illustrates the overall structure of MetaCD. The training process is as follows: we first initialize the network parameters $\theta_{M}$ and create a copy $\theta_{M^c}$. Then, response data are passed through the embedding module to obtain vectors $(\Tilde{S}_{ij}, \Tilde{Q}_{ij}, \Tilde{K}_{ij}, r_{ij})$. MetaCD randomly samples $m$ task units $\mathcal{T} = \{\mathcal{T}_1, \dots, \mathcal{T}_m\}$\footnote{$\mathcal{T}_i = t_r^{i}$; we use $\mathcal{T}_i$ here for clarity.} from $T_r$, and constructs their corresponding $\mathcal{Q}$-matrices. Using the support sets, it updates $\theta_{M^c}$ to obtain $\theta_{M^c}^{'}$. Next, the query sets are used to compute gradients from $\theta_{M^c}^{'}$, which are then used to update $\theta_M$, resulting in $\theta_M^*$. This optimized $\theta_M^*$ is used to initialize MetaCD for the test task $T_e$, and further fine-tuned using its support set. Finally, the model $M^*$ is obtained. Given the query set in $T_e$, MetaCD predicts each student's response (0 or 1).


\textbf{\textit{Embedding module:}}Let $A$, $B$, and $C$ denote the numbers of students, questions, and skills, with embedding dimension $d$ set to the number of skills per task. A $\mathcal{Q}$-matrix~\cite{qin2024exploration} maps questions to skills. Each response is encoded as $R_{ij} = (\Tilde{S}_{ij}, \Tilde{Q}_{ij}, \Tilde{K}_{ij}, r_{ij})$, where $\Tilde{S}_{ij}$, $\Tilde{Q}_{ij}$, and $\Tilde{K}_{ij}$ are one-hot vectors, and $r_{ij}$ is the score. Trainable embeddings estimate student proficiency, question difficulty, and discrimination. The knowledge base, initialized as $\theta_M$, is updated via task-specific copies $\theta_{M^c}$.

\textbf{\textit{Knowledge base module:}} Figure~1 illustrates the Knowledge Base (KB) module, comprising three components: KB train support, KB train query, and KB test support. The KB module uses meta-learning to accumulate and store experiential knowledge from multiple training task units. The base model $f(\theta_M)$ can be any deep model, such as MLP~\cite{ahakonye2024multi} or CNN~\cite{shah2023comprehensive}. The KB supports both updates and retrieval. The process is detailed as follows:

\textbf{1. KB Train Support:} We repeatedly sample tasks $\mathcal{T}$ from $T_r$, and use support sets $\mathcal{T}_s$ to train the KB with parameters $\theta_M$. Task units differ in dataset source, data distribution, etc. MetaCD learns from many $\mathcal{T}_s$ to store generalizable knowledge in KB. The specific expression is as follows:
    \begin{equation}
      \theta_{M^c} \leftarrow \theta_{M^c} - \alpha \cdot \frac{1}{n} \cdot \frac{1}{m} \sum_{i=1}^{n} \sum_{j=1}^{m} \nabla_{\theta_{M^c}}L(f(\theta_{M^c}),\mathcal{T}_s)
    \end{equation}
\noindent where \(n\) represents the number of \(\mathcal{T}_s\), which corresponds to the batch\_size, and \(L\) denotes the loss function. In this paper, we use the cross-entropy loss function.  we set the hyper-parameter $\alpha$ as 0.3. 

\textbf{2. KB Train Query:} To reduce redundancy, we refine $\theta_{M^c}$ using the query set $\mathcal{T}_q$ with a combined loss: $L_{total} = L(f(\theta_{M^c}),\mathcal{T}_q) + L_{PPM}$. The gradient from $L_{total}$ updates $\theta_M$, not $\theta_{M^c}$ (see Eq.~(2)). The updated $\theta_M$ is treated as the optimal initialization $\theta_M^*$. This step enhances the KB with query knowledge, improving task adaptation.  We set the hyper-parameter $\beta$ to 0.5.

\begin{equation}
    \begin{split}
        \nabla KB &= \nabla_{\theta_{M^c}} L_{\text{total}} \\
        \theta_M^* &= \theta_{M} \leftarrow \theta_{M} - \beta \cdot \nabla KB
    \end{split}
\end{equation}

\textbf{\textit{In KB Train Query}}, the gradient computed from \( L_{\text{total}} \) will no longer be used to update \( \theta_{M^c} \), but instead will be used to update \( \theta_{M} \) (see Eq. (2)). The parameters \( \theta_M \) obtained from KB train query are considered as the optimal initialization parameters \( \theta_M^* \). The process aims to optimize the parameters in the knowledge base using \( \mathcal{T}_q \), allowing MetaCD to directly leverage the knowledge from the knowledge base as initial knowledge when facing new or long-tailed tasks. This ensures that the model starts with a good knowledge initial state that includes prior knowledge, leading to better performance. 

\textbf{3. KB Test Support:} We initialize the knowledge base with $\theta_M^*$ and fine-tune it on the support set of the test task $T_e$. During this stage, we apply the parameter protection mechanism (see \textbf{\textit{Parameter protection module}}) to adapt KB for future tasks. After KB training, we remove the classification layer so that the knowledge base functions as a feature extractor.

\textbf{\textit{Parameter protection module:}} The Parameter Protection Mechanism (PPM) balances stability on old tasks and adaptability to new ones by integrating into the loss during KB train query and test support. It preserves key knowledge from training tasks and reduces forgetting across sequential tasks. PPM measures parameter sensitivity, restricting updates to sensitive parameters while allowing others to change freely. The calculation is shown in Eq.~(3).

\begin{equation}
    L_{PPM} = \frac{1}{2} \sum_{i,k} \phi_{i,k} (\theta_{i,k} - \theta_{i,k}^*)^2,\\
\end{equation}

\noindent where $\theta_{i,k}^*$ represents the $k$-th parameter obtained after training in the $i$-th task unit, $\theta_{i,k}$ indicates the $k$-th initialization value in the $i$-th task unit. $\phi_{i,k}$ refers to the sensitivity (importance) weights of the parameter, and its calculation formula is as follows:
\begin{equation}
   \phi_{i,k} = \frac{1}{N} \sum_{n=1}^{N} ||g_{ik}(x_n)||,
\end{equation}
\noindent where N refers to the total number of data in task unit, $g_{ik}$ represents the gradient of the KB with respect to the $x_n$, and $g_{ik}(x_n) = \frac{\partial KB(x_n, \theta)}{\partial \theta_{ik}}$. Here, KB stand for the output of the knowledge base module, and $\theta$ indicates the parameters of KB. The term $g_{ik}$ can be seen as the importance of $\theta_{i,k} $ at the data $x_n$. Consequently, $\phi_{i,k}$ can be seen as the importance of $k$-th parameter in the $i$-th task unit. 

\textbf{\textit{Per-class diagnosis module:}} In order to solve the problem of fuzzy boundaries, we adopt a Per-class-based diagnosis method. Since cognitive diagnosis results are binary (0 or 1), we build a separate fully connected network (FCN) for each class, called a per-class head. Each per-class head consists of 4 fully connected layers. The input to the per-class diagnosis module is the knowledge base output, and its output is computed as in $y=\underset{k}{\arg \max }\left(f_0\left(X_{KB}\right), f_1\left(X_{KB}\right)\right))$. Where $X_{KB}$ represents the output of knowledge base module, $f_0$ and $f_1$ indicates the two head of result 0 and result 1, respectively. The loss function of each head is shown in Eq. (5). 
\begin{equation}
\begin{split}
loss =\underset{\mathbf{X_{KB}} \sim \mathbb{P}_{\mathbf{X_{KB}}}}{\mathbb{E}}[-\log (\operatorname{Sigmoid}(f_i({X_{KB}})))]+ \\
\eta \cdot \underset{\mathbf{X_{KB}} \sim \mathbb{P}_{\mathbf{X_{KB}}}}{\mathbb{E}}\left\|\frac{\partial f_i(X_{KB})}{\partial X_{KB}}\right\|_2^{\mu} + \\
\lambda \cdot \left\| \theta_1 - \theta_{0}\right\|_2^2,
\end{split}
\end{equation}

\noindent where $X_{KB}$ represents the output of knowledge base module, $f_0$ and $f_1$ indicates the two head of result 0 and result 1, respectively. The first term in Eq.(5) is called the negative logarithmic function (NLL). We can make the output value of NLL as large as possible by reducing loss. However, we cannot make it infinitely large, otherwise it will cause severe overfitting, making the results between the two heads impossible to compare. Therefore, we introduce the holistic regularization (H-reg) as the second term of Eq.(5). This allows the parameters to be as large as possible while ensuring that the parameters and output are controllable. In this work, both heads are all 4-layer FCN with ReLU activation function, so H-reg can be expressed as Eq.(6). The third term in Eq.(5) is the L2-transfer regularization,  which aims to control the model complexity by penalizing large weight values. Moreover, in order to make the output values of $f_1$ and $f_0$ be in the same value space as much as possible, the parameters of $f_1$ is initialized with $f_0$.
\begin{equation}
\underset{\mathbf{X_{KB}} \sim \mathbb{P}_{\mathbf{X_{KB}}}}{\mathbb{E}}\left\|\frac{\partial f_i(X_{KB})}{\partial X_{KB}}\right\|_2^{\mu} = \underset{\mathbf{X_{KB}} \sim \mathbb{P}_{\mathbf{X_{KB}}}}{\mathbb{E}}\left\| \omega_4 \cdot \omega_3 \cdot \omega_2 \cdot \omega_1 \right\|_2^{\mu},
\end{equation}

To address the existence of fuzzy boundaries, after training is completed, we adopt a method based on Kullback-Leibler divergence \cite{yang2023active} to reduce the shared knowledge, thereby increasing the separation distance between different diagnosis result. The calculation formula for knowledge sharing between different diagnosis results is shown in Eq. (7).
\begin{equation}
\Lambda^* =\operatorname{argmin} \sum_{i=0}^1 \kappa_i KL\left(\mathbb{P}_i \| \mathbb{P}_{0: 1}\right),
\end{equation}

\section{Experimental analysis and results}
\label{s4}

\textbf{\textit{Comparison Baselines:}} we conducted a comparative analysis with several existing cognitive diagnosis baselines using the aforementioned dataset. These baseline methods mainly include traditional cognitive diagnostic models represented by IRT \cite{edelen2007applying}, MIRT \cite{chalmers2012mirt}, DINA \cite{de2009dina} and BCD \cite{bi2023beta}, and neural network-based cognitive diagnostic models represented by NCD \cite{wang2020neural}, RCD \cite{gao2021rcd}, and SCD \cite{wang2023self}.

\noindent We use the ASSIST (ASSIST2009\_2010~\cite{wang2020neural}, ASSIST2012\_2013~\cite{liu2023homogeneous}, ASSIST2017~\cite{wang2023self}), CDBD\_a0910, and NIPS2020 datasets (as detailed in \hyperref[sec:dataset]{Appendices A.1}) to evaluate MetaCD on the following research questions: \textbf{\textit{RQ1}}: Can MetaCD achieve high accuracy in cognitive diagnosis compared to baseline models? \textbf{\textit{RQ2}}: Can MetaCD maintain strong generalization under long-tailed data distributions? \textbf{\textit{RQ3}}: Can MetaCD remain stable and adaptable in task-incremental learning? \textbf{\textit{RQ4}}: What is the impact of each module through ablation study? \hyperref[sec:setup]{Appendices A.2} shows the detailed experimental setup.

\begin{table*}
\caption{Performance comparison of different models on student cognitive diagnostic. }
\centering

\setlength{\tabcolsep}{3mm}{
\begin{tabular}{lccccccc}
\hline
\\[-2ex] 
\textbf{Work} & \multicolumn{3}{l}{ASSIST2009\_2010} & \multicolumn{1}{l}{} &\multicolumn{3}{l}{ASSIST2017} \\
\cline{2-4} \cline{6-8} 
\\[-4.5ex] 
\\
      & ACC   & RMSE  & AUC &   & ACC   & RMSE  & AUC   \\  
\hline
\textbf{MetaCD}  & \textbf{0.753} & 0.425 & \textbf{0.771}&  & \textbf{0.715} & \textbf{0.439} & \textbf{0.726} \\
IRT    & 0.654 & 0.472 & 0.681&  & 0.658 & 0.464 & 0.668 \\
MIRT  & 0.707 & 0.461 & 0.716&  & 0.668 & 0.461 & 0.678 \\
DINA  & 0.644 & 0.495 & 0.680&  & 0.613 & 0.519 & 0.654 \\
BCD  & 0.729 & 0.426 & 0.763&  & 0.701 & 0.447 & 0.713 \\
NCD & 0.726 & 0.441 & 0.752&  & 0.685 & 0.453 & 0.699 \\
RCD & 0.724 & 0.427 & 0.761&  & 0.694 & 0.450 & 0.709\\ 
SCD  & 0.731 & \textbf{0.423} & 0.729&  & 0.703 & 0.442 & 0.710 \\
\hline
\end{tabular}
}
\label{tab1}
\end{table*}

\textbf{\textit{RQ.1: }} \hyperref[tab1]{Table~1} shows that MetaCD achieves the best ACC and AUC, with slightly higher RMSE than RCD, suggesting better task initialization. Considering data privacy issues, we further test MetaCD in few-shot settings. \hyperref[tab2]{Table~2} and \hyperref[tab2]{Table~3} ( in \hyperref[sec:experiment]{Appendices A.4}) show that as samples increase, all models improve, but MetaCD performs best, showing strong generalization and robustness under limited data.

\textbf{\textit{RQ.2:}} As NIPS2020 is used for meta-training and is uniformly distributed, we exclude it from testing. We evaluate MetaCD on ASSIST2009\_2010, ASSIST2012\_2013, ASSIST2017, and CDBD\_a0910. For RQ2, we group interactions by frequency (6--10 to 31--35). \hyperref[fig2]{Figure~2} shows MetaCD consistently outperforms SCD, showing strong generalization on sparse data.

\textbf{\textit{RQ.3:}} To answer RQ3, we evaluate MetaCD in a task-incremental setting. Trained sequentially on four datasets after NIPS2020, MetaCD with parameter protection achieves a BWT ( in \hyperref[sec:metrics]{Appendices A.3}) of -0.04 vs. -0.217 without it \hyperref[tab4]{Table~4}} ( in \hyperref[sec:experiment]{Appendices A.4}), showing reduced forgetting and better stability.

\textbf{\textit{RQ.4:}} To evaluate the contribution of each module in MetaCD, we conduct ablation experiments on four datasets. As shown in \hyperref[tab5]{Table~4} ( in \hyperref[sec:experiment]{Appendices A.5}), removing any module degrades performance. Notably, removing the KB module results in the largest accuracy drop (average -3.65\%), highlighting its core role. Excluding the PPM module reduces ACC by 1.8\% on average, showing its effectiveness in filtering redundant knowledge during KB query and support stages. The absence of the Per-class module also harms performance, confirming its utility in resolving fuzzy class boundaries.

\section{Conclusion and Limitations}
\noindent This article addresses the challenges posed by long-tailed problems and dynamic changes in cognitive diagnosis, offering a meta-learning framework based on continual learning as a solution.  Specifically, we first set a meta-learning-based task objective to enable the neural network to mine prior knowledge from past tasks, thereby better generalizing on the long-tailed data. Then, we use a continual learning method based on parameter protection mechanism to calculate the importance level of parameters so that the model adapts well to new tasks and generalizes well to new skills. This ensures the stability and plasticity of the model. Although MetaCD outperforms existing baseline methods, computational efficiency has become an important consideration. Specifically, in PPM, the need to compute the sensitivity of each parameter adds additional computational overhead. Future work will focus on improving the computational efficiency of MetaCD.

\bibliographystyle{ieeetr}
\bibliography{custom}

\newpage
\appendix
\section{Technical Appendices}

\subsection{Dataset Description}
\label{sec:dataset}
In our experiments, we employ the following publicly available datasets: ASSIST,  CDBD$\_$a0910 and NIPS2020. ASSIST, comprising ASSIST2009$\_$2010 \cite{wang2020neural}, ASSIST2012$\_$2013 \cite{liu2023homogeneous}, and ASSIST2017 \cite{wang2023self}, is an openly accessible dataset sourced from the ASSISTments online learning platform \cite{feng2023promising}. This dataset encompasses students' responses to mathematics questions. CDBD$\_$a0910 stands as one of the benchmark datasets in cognitive diagnosis, offering clearly student answer logs alongside corresponding information on knowledge concepts pertaining to the each exercise \footnote{https://github.com/bigdata-ustc/EduData.}. NIPS2020 is supported by the NeurIPS 2020 Education Challenge competition, offering answer logs from students regarding multiple-choice math questions \cite{liu2023homogeneous}. Type indicates the storage format of the original data. 

The $\#$ of Records  represents the number of response logs. The $\#$ of Student, $\#$ of Exercises and $\#$ of Knowledge concepts refer to the number of students, exercises, and knowledge concepts, respectively. Due to the presence of null records, missing values, and duplicate entries in the original datasets, preprocessing is required for all datasets before conducting the experiment. In addition, to demonstrate the performance of SCD on long-tailed data, we retain students' data with more than five interaction records. The preprocessed data is stored as files in JSON format.

\subsection{Experimental Setup}
\label{sec:setup}

In this experiment, we have set the total$\_$tasks to 50, meaning that we have prepared 50 task units for training MetaCD. The details are outlined as follows: (1) ASSIST2009$\_$2010, ASSIST2012$\_$2013, and ASSIST2017 are treated as separate task units. (2) Considering that the CDBD$\_$a0910 original data comprises three sub-datasets: train.csv, test.csv, and valid.csv, and each sub-dataset contains a specific number of records, we divide the CDBD$\_$a0910 dataset into 3 task units. (3) Given the large volume of records in NIPS2020, we divide it into 115 task units, each containing 150,000 records. 

From these, we randomly select 45 task units for model training (m = 45). In MetaCD, the selection of hyperparameters is primarily optimized using a grid search method. Specifically, the values of $\alpha$, $\beta$, $\eta$, $\lambda$, and $\xi$ are in the range of (0, 1], with a step size of 0.1 for each change. The value of $\mu$ is selected from the set [2, 3, 4]. Furthermore, we set batch\_size = 5, and the number of sampling samples $s$ = 128. This implies that during each epoch of KB train support, we will select 5 training task units and sample 128 data points for each task unit to train the model.

In addition, we use Xavier initialization to initialize the MetaCD, aiming to speed up training and improve the performance of the network; and in MetaCD, we use the Dropout algorithm (dropout rate = 0.5) to reduce the problem of overfitting in the model process and use Adam The optimization algorithm performs gradient updates to help the model quickly converge to the optimal solution.


\noindent \textbf{Problem Definition:} Given a set of students' response data from training task units set $T_r$ and $\mathcal{Q}$-matrix, the purpose of MetaCD is to learn a model that can predict students' proficiency in skills and their corresponding scores by effectively transferring knowledge across tasks. The ultimate goal is to obtain the best model that can accurately predict students' performance on the testing task $T_e$. 

\subsection{Evaluation metrics}
\label{sec:metrics}
In order to better evaluate the performance of MetaCD, we used overall accuracy (ACC), root mean square error (RMSE) and area under the curve (AUC) to evaluate the performance of MetaCD from different perspectives. In addition, in order to evaluate the continual learning ability of MetaCD on task incremental learning, we introduce the backward transfer (BWT) to evaluate the model. The BWT is an evaluation metric utilized to assess the performance of deep learning models in task incremental learning scenarios. It primarily involves assessing the model's performance across all tasks, including both trained and untrained ones, subsequent to training it on a specific task.
\begin{equation}
	BWT = \frac{1}{T-1} \sum_{t=1}^{T-1}{M_{T,t} - M_{t,t}}
\end{equation}
\noindent where R $\in$ (1, T-1), $M_{R,i}$ indicates the performance of the model on the task $t$ after the model is trained on the $R^{th}$ task.

\subsection{Detailed Results of Experiments}
\label{sec:experiment}

\begin{figure}
    \centering
    \includegraphics[width=1.0\textwidth]{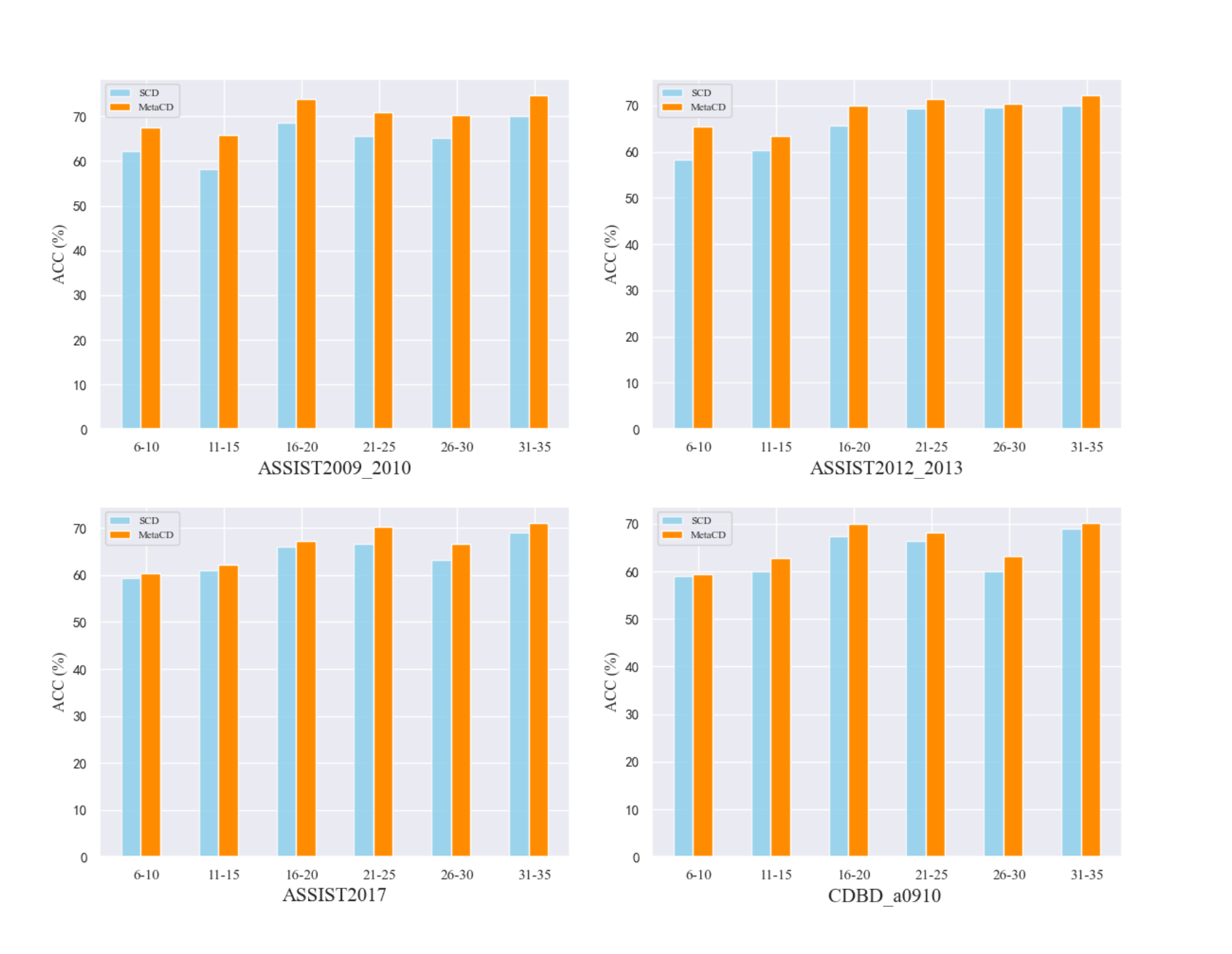}
    \caption{Performance comparison on long-tailed data. The horizontal coordinate represents the number of times that a question has been answered by different students. For example, 5-10 represents a question that has been answered by different students between 5 and 10 times.}
    \label{fig2}
\end{figure}

\begin{table}
\centering
\caption{Performance comparison of models based on the ASSIST 2009\_2010 under different data amounts.}
\resizebox{\linewidth}{!}{
\setlength{\tabcolsep}{3mm}{
\begin{tabular}{c|ccc|ccc|ccc|ccc}
\hline
\scriptsize Amounts &\multicolumn{3}{c|}{5,000} &\multicolumn{3}{c|}{100,000} &\multicolumn{3}{c|}{150,000} &\multicolumn{3}{c}{190,000} \\ 
\cline{2-13}
\hline
\scriptsize Work &\scriptsize ACC &\scriptsize RMSE &\scriptsize AUC &\scriptsize ACC &\scriptsize RMSE &\scriptsize AUC &\scriptsize ACC &\scriptsize RMSE &\scriptsize AUC &\scriptsize ACC &\scriptsize RMSE &\scriptsize AUC\\
\hline
\scriptsize \textbf{MetaCD} &\scriptsize \textbf{0.569} &\scriptsize \textbf{0.517}&\scriptsize \textbf{0.557 }&\scriptsize \textbf{0.659}&\scriptsize \textbf{0.483}&\scriptsize \textbf{0.687} &\scriptsize \textbf{0.738}&\scriptsize \textbf{0.459} &\scriptsize \textbf{0.724}&\scriptsize \textbf{0.753} &\scriptsize 0.425&\scriptsize \textbf{0.771} \\
\scriptsize IRT &\scriptsize 0.463 &\scriptsize 0.693&\scriptsize 0.419 &\scriptsize 0.532&\scriptsize 0.690&\scriptsize 0.526 &\scriptsize 0.585&\scriptsize 0.591 &\scriptsize 0.613&\scriptsize 0.654 &\scriptsize 0.472&\scriptsize 0.681 \\
\scriptsize MIRT &\scriptsize 0.469 &\scriptsize 0.690&\scriptsize 0.428 &\scriptsize 0.549&\scriptsize 0.675&\scriptsize 0.568 &\scriptsize 0.605&\scriptsize 0.570 &\scriptsize 0.628&\scriptsize 0.707 &\scriptsize 0.461&\scriptsize 0.716 \\
\scriptsize DINA &\scriptsize 0.476 &\scriptsize 0.667&\scriptsize 0.462 &\scriptsize 0.596&\scriptsize 0.607&\scriptsize 0.585 &\scriptsize 0.651&\scriptsize 0.519 &\scriptsize 0.633&\scriptsize 0.644 &\scriptsize 0.495&\scriptsize 0.680 \\
\scriptsize BCD &\scriptsize 0.556 &\scriptsize 0.572 &\scriptsize 0.539 &\scriptsize 0.621 &\scriptsize 0.549 &\scriptsize 0.618  &\scriptsize 0.722 &\scriptsize 0.473  &\scriptsize 0.706  &\scriptsize 0.729 &\scriptsize 0.426&\scriptsize 0.763 \\
\scriptsize NCD &\scriptsize 0.503 &\scriptsize 0.581&\scriptsize 0.529 &\scriptsize 0.628&\scriptsize 0.536&\scriptsize 0.629 &\scriptsize 0.719&\scriptsize 0.493 &\scriptsize 0.703&\scriptsize 0.726 &\scriptsize 0.441&\scriptsize 0.752 \\
\scriptsize RCD &\scriptsize 0.557 &\scriptsize 0.569&\scriptsize 0.535 &\scriptsize 0.615 &\scriptsize 0.542&\scriptsize  0.626 &\scriptsize 0.705 &\scriptsize 0.479  &\scriptsize 0.698&\scriptsize 0.724 &\scriptsize 0.427&\scriptsize 0.761\\
\scriptsize SCD &\scriptsize 0.560 &\scriptsize 0.522&\scriptsize 0.542 &\scriptsize 0.637&\scriptsize 0.505&\scriptsize 0.679 &\scriptsize 0.725&\scriptsize 0.466 &\scriptsize 0.709&\scriptsize 0.731 &\scriptsize \textbf{0.423}&\scriptsize 0.729 \\
\hline
\end{tabular}}}
\label{tab2}
\end{table}

\begin{table*}
\centering
\caption{Performance comparison of models based on the ASSIST2017 under different data amounts.}
\resizebox{\linewidth}{!}{
\setlength{\tabcolsep}{3mm}{
\begin{tabular}{c|ccc|ccc|ccc|ccc}
\hline
\scriptsize Amounts &\multicolumn{3}{c|}{5,000} &\multicolumn{3}{c|}{10,0000} &\multicolumn{3}{c|}{15,0000} &\multicolumn{3}{c}{19,0000} \\ 
\cline{2-13}
\hline
Work &\scriptsize ACC &\scriptsize RMSE &\scriptsize AUC &\scriptsize ACC &\scriptsize RMSE &\scriptsize AUC &\scriptsize ACC &\scriptsize RMSE &\scriptsize AUC &\scriptsize ACC &\scriptsize RMSE &\scriptsize AUC\\
\hline
\scriptsize \textbf{MetaCD} &\scriptsize \textbf{0.553} &\scriptsize \textbf{0.550}&\scriptsize \textbf{0.549} &\scriptsize \textbf{0.646}&\scriptsize \textbf{0.499}&\scriptsize \textbf{0.652} &\scriptsize \textbf{0.703}&\scriptsize \textbf{0.465} &\scriptsize \textbf{0.715}&\scriptsize \textbf{0.715} &\scriptsize \textbf{0.439}&\scriptsize \textbf{0.726} \\
\scriptsize IRT &\scriptsize 0.458 &\scriptsize 0.685&\scriptsize 0.425 &\scriptsize 0.526&\scriptsize 0.697&\scriptsize 0.518 &\scriptsize 0.580&\scriptsize 0.599 &\scriptsize 0.601&\scriptsize 0.658 &\scriptsize 0.464&\scriptsize 0.668\\
\scriptsize MIRT &\scriptsize 0.483 &\scriptsize 0.679 &\scriptsize 0.417 &\scriptsize 0.535&\scriptsize 0.683&\scriptsize 0.547 &\scriptsize 0.576&\scriptsize 0.586 &\scriptsize 0.618&\scriptsize 0.668 &\scriptsize 0.461&\scriptsize 0.678 \\
\scriptsize DINA &\scriptsize 0.459  &\scriptsize 0.665&\scriptsize 0.453 &\scriptsize  0.553&\scriptsize 0.639&\scriptsize 0.571 &\scriptsize0.609&\scriptsize 0.558 &\scriptsize 0.607&\scriptsize 0.613 &\scriptsize 0.519&\scriptsize 0.654 \\
\scriptsize BCD &\scriptsize 0.540   &\scriptsize 0.565 &\scriptsize 0.528  &\scriptsize 0.635   &\scriptsize 0.510  &\scriptsize 0.645  &\scriptsize 0.690 &\scriptsize 0.513  &\scriptsize 0.692 &\scriptsize 0.701 &\scriptsize 0.447&\scriptsize 0.713 \\
\scriptsize NCD &\scriptsize 0.523 &\scriptsize 0.572&\scriptsize 0.528 &\scriptsize 0.609&\scriptsize 0.551&\scriptsize 0.603 &\scriptsize 0.689&\scriptsize 0.537 &\scriptsize 0.673&\scriptsize 0.685 &\scriptsize 0.453&\scriptsize 0.699 \\
\scriptsize RCD &\scriptsize 0.537 &\scriptsize 0.570 &\scriptsize 0.530 &\scriptsize 0.626 &\scriptsize 0.532&\scriptsize 0.638 &\scriptsize 0.685 &\scriptsize 0.519 &\scriptsize 0.682&\scriptsize 0.694 &\scriptsize 0.450&\scriptsize 0.709\\
\scriptsize SCD &\scriptsize 0.541 &\scriptsize 0.562 &\scriptsize 0.531 &\scriptsize 0.628&\scriptsize 0.515&\scriptsize 0.642 &\scriptsize 0.692&\scriptsize 0.489 &\scriptsize 0.703&\scriptsize 0.703 &\scriptsize 0.442&\scriptsize 0.710 \\
\hline
\end{tabular}}}
\label{tab3}
\end{table*}

\begin{table}
~\\
\centering
\makeatletter\def\@captype{table}\makeatother\caption{The performance of MetaCD on task incremental learning across sequential tasks.}
\begin{tabular}{ccccc} 
    		\toprule[0.1pt]  
     	      \rule{0pt}{10.3pt} & $Task_1$ & $Task_2$ & $Task_3$ & $Task_4$ \\
		\midrule[0.1pt]
		 $T_{1}^{*}$ $|$ $T_{1}$ & 0.771 $/$ 0.771&  /   &  / &  / \\
		 $T_{2}^{*}$ $|$ $T_{2}$ & 0.719 $/$ 0.598 & 0.703 $/$ 0.721 & /  & / \\
		 $T_{3}^{*}$ $|$ $T_{3}$ & 0.706 $/$ 0.533 & 0.689 $/$ 0.557& 0.700 $/$ 0.715 & /\\
          $T_{4}^{*}$ $|$ $T_{4}$ & 0.693 $/$ 0.506 & 0.675 $/$ 0.531 & 0.686 $/$ 0.519 & 0.697 $/$ 0.701 \\
		\bottomrule[0.1pt]   
    	  \end{tabular}
\label{tab4}
\end{table}

\begin{table}
~\\
\centering
\fontsize{10}{12}\selectfont
 \makeatletter\def\@captype{table}\makeatother\caption{Accuracy results of ablation study with/without KB, PPM, and Per-class on four different datasets.}
\setlength{\tabcolsep}{3mm}{
\begin{tabular}{ccccc}
\toprule[0.1pt]  
Model & ASSIST2009\_2010 & ASSIST2012\_2013 & ASSIST2017 & CDBD\_a0910 \\
\midrule[0.1pt]
MetaCD & 0.753 & 0.725 & 0.715 & 0.712 \\
w/o KB & 0.719 & 0.685 & 0.687 & 0.668 \\
w/o PPM & 0.733 & 0.708 & 0.699 & 0.693 \\
w/o Per-class & 0.742 & 0.713 & 0.706 & 0.705 \\
\bottomrule[0.1pt]
\end{tabular}
}
\label{tab5}
\end{table}

\end{document}